\documentclass[10pt]{article} % For LaTeX2e
\usepackage[preprint]{tmlr}

\usepackage{amsmath,amsfonts,bm}

\def\eqref#1{equation~\ref{#1}}
\def\1{\bm{1}}

\DeclareMathAlphabet{\mathsfit}{\encodingdefault}{\sfdefault}{m}{sl}
\SetMathAlphabet{\mathsfit}{bold}{\encodingdefault}{\sfdefault}{bx}{n}

\usepackage{amsmath,amssymb}
\usepackage{booktabs}
\usepackage{flafter}
\usepackage{array}
\usepackage{float}
\usepackage{graphicx}
\usepackage{microtype}
\usepackage{multirow}
\usepackage{xcolor}
\usepackage{colortbl}
\usepackage{hyperref}
\usepackage{url}
\hypersetup{hidelinks,pdfauthor={},pdftitle={ATLAS: Attention-Guided Test-Time Latent Selection for Compositional Visual Autoregressive Generation}}

\title{ATLAS: Attention-Guided Test-Time Latent Selection \\for Compositional Visual Autoregressive Generation}

\author{%
Hossein Shahabadi \quad
Raya Namazi \quad
Mahdieh Soleymani Baghshah\\[0.5em]
Sharif University of Technology\\[0.1em]
\texttt{hosseinshahabadi.me@gmail.com}
}

\def\month{MM}  % Insert correct month for camera-ready version
\def\year{YYYY} % Insert correct year for camera-ready version
\def\openreview{\url{https://openreview.net/forum?id=XXXX}} % Insert correct link to OpenReview for camera-ready version

\newcommand{\method}{ATLAS}
\definecolor{VISTABlue}{HTML}{245F9B}
\definecolor{BlueHeaderTint}{HTML}{E3EFF8}
\definecolor{BlueStripe}{HTML}{F1F6FB}
\definecolor{ATLASBlueTint}{HTML}{D9EAF6}
\definecolor{SelectedColumnTint}{HTML}{EAF2F9}
\newcommand{\Infinity}{Infinity}
\newcommand{\loss}{\mathcal{L}}
\newcommand{\attn}{\mathbf{A}}
\newcommand{\cand}{\mathcal{K}}

\begin{document}
\raggedbottom
\maketitle

\begin{abstract}
Visual autoregressive models produce convincing images but can misassign attributes and spatial relations. Searching multiple generation paths can improve composition, yet completing every candidate is costly. We introduce \method{}, a training-free framework that uses the frozen generator's internal attention to guide search before final images are available. Within-prompt diagnostics identify attention-derived scores that predict final compositional preferences on \Infinity{}, with category-dependent transfer to Switti. \method{} uses these on-the-fly signals to progressively concentrate computation on promising trajectories. It maintains multiple coarse alternatives, prunes candidates without decoding at intermediate scales, and reserves image-space evaluation for a final choice among partial trajectories. \method{} scales effectively on both backbones, achieving higher mean scores on the evaluated T2I-CompBench and GenEval categories as more computation is devoted to search. At the largest tested budget, the targeted T2I-CompBench mean improves by over $10\%$ relative to each backbone. Separate profiles show over $50\%$ less generation time than completing and evaluating all candidates at equal initial counts, alongside lower peak allocated GPU memory. Relative time savings are larger at the higher tested candidate count. Together, these results support scalable compositional search: additional sampling expands exploration, while attention-guided selection limits continuation cost without parameter updates or gradient-based optimization.
\end{abstract}

\section{Introduction}
\label{sec:introduction}

Diffusion models generate images by iterative denoising~\citep{rombach2022latentdiffusion}; visual autoregressive (VAR) models predict tokens across progressively finer scales~\citep{tian2024var}. Infinity extends next-scale prediction to high-resolution text-to-image synthesis, while Switti uses a distinct scale-wise transformer design~\citep{han2025infinity,voronov2025switti}. Both expose intermediate states that can be inspected before generation is complete.

High visual quality does not ensure correct composition. An image may misassign an attribute, omit an entity, or reverse a relation. T2I-CompBench++ and GenEval evaluate such errors~\citep{huang2025t2icompbenchpp,ghosh2023geneval,shahabadi2025infinitybeyond}. Sampling alternatives creates opportunities for better composition, but completing every alternative incurs the full generation cost. Early selection avoids some of that work only if an intermediate score identifies paths worth continuing.

Test-time interventions address composition through steering or search. VISTA steers intermediate representations with attention constraints, and SynVAR guides spatial and semantic generation~\citep{shahabadi2026vista,chen2026synvar}. TTS-VAR preserves coarse diversity before potential-based path selection~\citep{chen2025ttsvar}; related next-token and raster-scan methods use token confidence and progressive grid partitions~\citep{chen2026scalingar,park2026gridar}. Attention offers another selection signal by connecting prompt tokens to visual positions without decoding candidates. Its use for steering does not establish whether it predicts which sampled trajectory will finish better.

We ask whether a frozen generator's attention can allocate search computation toward better compositional outcomes. We choose losses and layers on calibration prompts, then test their predictions among held-out candidates for the same prompt. Infinity scores achieve 72.9\% pairwise agreement for planar relations and 62.7\% for color binding over scale indices 2--6, against 50\% chance. Selected color and texture losses also rank Switti candidates above chance by their prompt-bootstrap intervals. Scale and layer analyses characterize signal availability; a separate prompt-switch experiment provides context on early visual commitments.

Attention-Guided Test-Time Latent Selection (ATLAS) turns intermediate scores into a search policy. It retains several coarse candidates, applies three attention-guided cuts, and decodes the remaining partial images for one ImageReward decision~\citep{xu2023imagereward}. Only one survivor completes the finer scales. This division uses internal scores for repeated selection and reserves image-space evaluation for the final commitment. The generator remains frozen, with no parameter updates or gradient-based latent optimization.

To our knowledge, ATLAS is the first attention-guided compositional search framework evaluated across two distinct VAR model families. End-to-end T2I-CompBench and GenEval averages increase with search budget on Infinity and Switti. Separate profiles with $8$ or $16$ initial candidates show $52$--$73\%$ less generation time than completing and evaluating the same number, with lower peak allocated memory. These results assess the complete controller; the diagnostic study separately establishes the predictive value of its attention-based scores.

Our contributions are as follows:
\begin{itemize}
  \item We study when prompt changes affect next-scale generation and how attention-derived compositional signals vary with scale, layer selection, and constraint category.
  \item We evaluate those signals through within-prompt final-image preferences, freeze loss and layer choices before held-out testing, and test their transfer from Infinity to Switti.
  \item We introduce ATLAS, a training-free controller that combines internal attention pruning with one image-space selection stage, and evaluate its compositional scores across search budgets alongside generation time and GPU memory.
\end{itemize}

\section{Preliminaries}
\label{sec:preliminaries}

Early selection compares unfinished trajectories, not completed images. We first define what a trajectory contains and how its attention yields a score available at a scale boundary.

\paragraph{Next-scale generation yields a sequence of candidate prefixes.}
VAR predicts each visual scale conditional on the maps already sampled~\citep{tian2024var}. This lets a search controller retain a prefix and generate only its continuation. Let $p$ be a text prompt, $c$ its text-conditioning sequence, and $r_s$ the token map at scale $s$. For $S$ scales, the corresponding factorization is
\begin{equation}
  P(r_{0:S-1}\mid c)
  = \prod_{s=0}^{S-1} P(r_s\mid r_{<s},c).
  \label{eq:next-scale}
\end{equation}
Earlier maps form input $z_s\in\mathbb{R}^{L_s\times C}$, with $L_s=h_sw_s$ visual positions and feature dimension $C$. Infinity uses bitwise prediction; Switti has a distinct scale-wise design~\citep{han2025infinity,voronov2025switti}. Our configurations have 13 scales and 32 layers for Infinity, and 14 scales and 30 recorded layers for Switti. Scales are zero-based; layers are one-based. Candidate $k$ has prefix $r_{k,\leq s}$ through scale $s$ and completed image $x_k$.

\paragraph{Cross-attention connects visual positions to prompt tokens.}
Cross-attention is a candidate diagnostic because it connects visual positions to prompt tokens. These maps support diffusion editing and compositional steering~\citep{hertz2022prompttoprompt,chefer2023attend,agarwal2023astar,li2023divide,rassin2023syngen}. VISTA uses their constraints for gradient descent on intermediate representations of a frozen next-scale generator~\citep{shahabadi2026vista}. We denote candidate $k$'s map at scale $s$ and layer $l$ by $\attn_{k,s}^{(l)}$. ATLAS reads it without modifying the trajectory.

\paragraph{Attention-derived objectives provide candidate scores.}
To compare candidates, a map must be reduced to a prompt-dependent scalar. Prior objectives measure attribute--noun agreement or separation between entity maps~\citep{rassin2023syngen,meral2024conform,shahabadi2026vista}. Let $\phi_d(A,c)$ be such an objective for category $d$, oriented so that lower values predict better composition. A layer set $W_d$ specifies which maps the objective reads:
\begin{equation}
  \bar{\attn}_{k,s}^{W_d}
  = \frac{1}{|W_d|}\sum_{l\in W_d}\attn_{k,s}^{(l)},
  \qquad
  \loss_d(k,s;W_d)
  = \phi_d\!\left(\bar{\attn}_{k,s}^{W_d},c\right).
  \label{eq:loss}
\end{equation}
Comparing $\loss_d(i,s;W_d)$ and $\loss_d(j,s;W_d)$ orders two candidates after the same scale. The definition includes sign conventions and averages maps \emph{before} evaluating the objective, rather than averaging nonlinear layer losses. Table~\ref{tab:atlas-frozen-scores} lists the chosen objectives; Appendix~\ref{app:loss-formulas} explains how they are computed. We test whether the lower-scoring candidate produces a final image that better satisfies the prompt.

\section{What Does Attention Reveal Before an Image Is Complete?}
\label{sec:attention}

An intermediate score is useful only if it distinguishes continuations worth computing. We examine this requirement in two ways. Prompt switches probe whether early prefixes constrain the eventual image. Completed same-prompt candidates then provide a direct test of attention-based prediction, including its dependence on scale and layer choice.

\subsection{Prompt sensitivity declines at different scales for different properties}

How strongly does an early prefix constrain its continuation? We replace the prompt while retaining sampled visual tokens. Unlike its use for diffusion editing~\citep{hertz2022prompttoprompt}, replacement here probes responsiveness. With $c_A,c_B$ encoding the original and replacement prompts, a switch after completed scale $s$ changes the condition at scale $u$ as follows:
\begin{equation}
  r_u^{A\to B(s)}
  \sim P\!\left(r_u\mid r_{<u}^{A\to B(s)},c_u^{(s)}\right),
  \qquad
  c_u^{(s)}=
  \begin{cases}
    c_A,&u\leq s,\\
    c_B,&u>s.
  \end{cases}
  \label{eq:prompt-switch}
\end{equation}
We save the prefix and its continuation state, re-encode B, and resume at scale $s+1$ without replacing earlier visual tokens. Indices start at zero: switching after scale 2 keeps the first three scales generated under A and uses B from scale 3 onward.

\begin{figure}[t]
  \centering
  \includegraphics[width=0.70\linewidth]{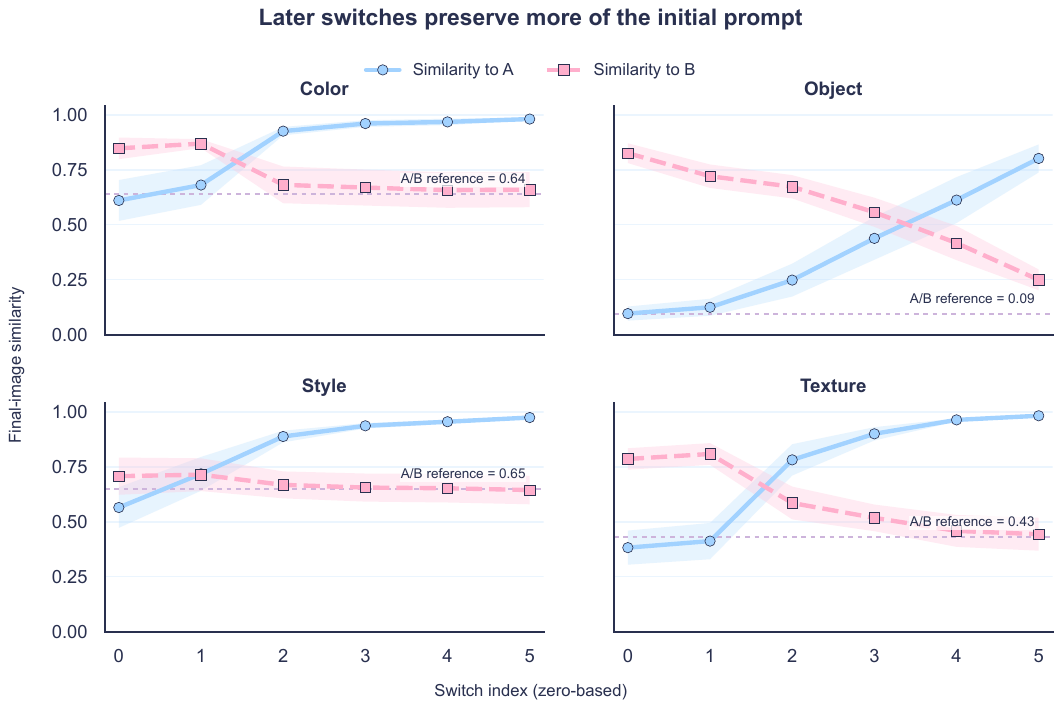}
  \caption{Final-image similarity after replacing A with B. A horizontal-axis value $s$ means scale $s$ has been generated under A before replacement; B is used from scale $s+1$ onward (Equation~\ref{eq:prompt-switch}). Blue/pink curves show image-embedding cosine similarity to full A/B references generated with the same seed. Dotted lines show reference similarity. Each point covers ten prompt pairs. Appendix~\ref{app:switch-code} details the scoring protocol and the scope of these descriptive summaries.}
  \label{fig:swap}
\end{figure}

\begin{figure*}[t]
\centering
\begin{minipage}{0.80\textwidth}
\centering
\setlength{\tabcolsep}{1pt}
\renewcommand{\arraystretch}{0.4}
\begin{tabular}{@{}c@{\hspace{2pt}}*{7}{>{\centering\arraybackslash}p{0.13\linewidth}}@{}}
& {\fontsize{7pt}{7.2pt}\selectfont Prompt A}
& {\fontsize{7pt}{7.2pt}\selectfont s4}
& {\fontsize{7pt}{7.2pt}\selectfont s3}
& {\fontsize{7pt}{7.2pt}\selectfont s2}
& {\fontsize{7pt}{7.2pt}\selectfont s1}
& {\fontsize{7pt}{7.2pt}\selectfont s0}
& {\fontsize{7pt}{7.2pt}\selectfont Prompt B}
\\[1pt]
\multicolumn{8}{@{}l}{\fontsize{7pt}{7pt}\selectfont\textbf{Color}}\\[1pt]
\rotatebox{90}{\scriptsize\textbf{Fish}}
& \includegraphics[width=\linewidth]{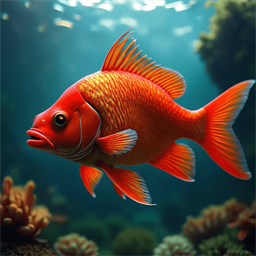}
& \includegraphics[width=\linewidth]{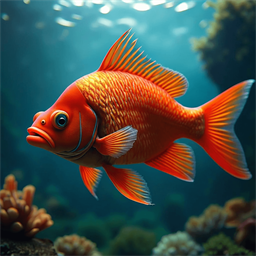}
& \includegraphics[width=\linewidth]{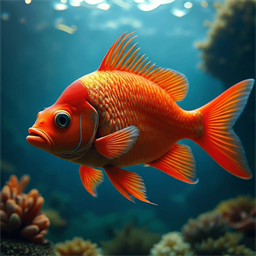}
& \includegraphics[width=\linewidth]{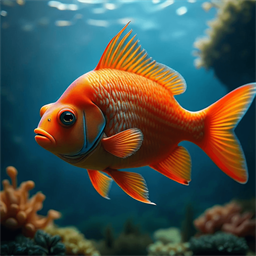}
& \includegraphics[width=\linewidth]{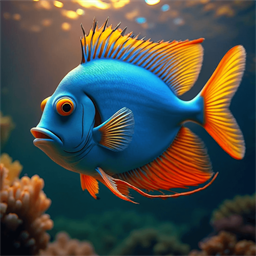}
& \includegraphics[width=\linewidth]{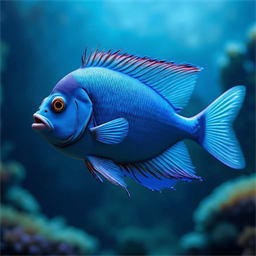}
& \includegraphics[width=\linewidth]{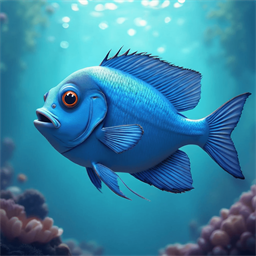}
\\[1pt]
\rotatebox{90}{\scriptsize\textbf{House}}
& \includegraphics[width=\linewidth]{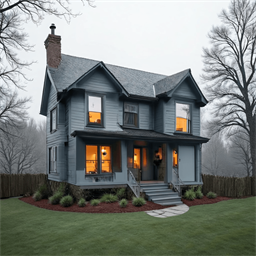}
& \includegraphics[width=\linewidth]{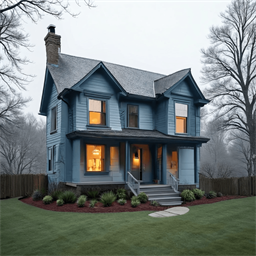}
& \includegraphics[width=\linewidth]{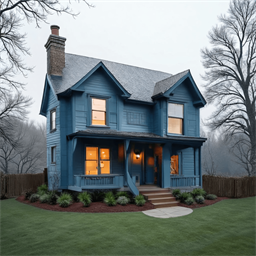}
& \includegraphics[width=\linewidth]{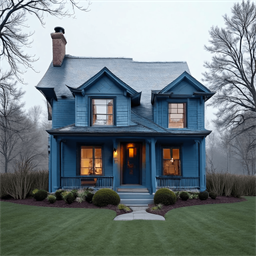}
& \includegraphics[width=\linewidth]{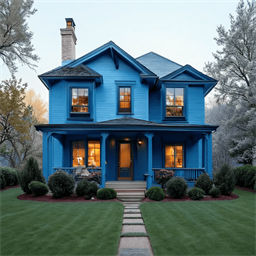}
& \includegraphics[width=\linewidth]{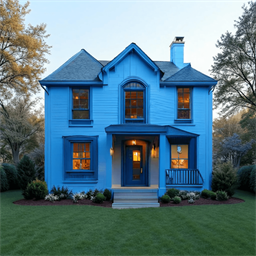}
& \includegraphics[width=\linewidth]{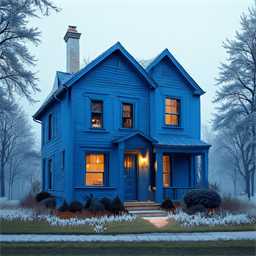}
\\[3pt]
\multicolumn{8}{@{}l}{\fontsize{7pt}{7pt}\selectfont\textbf{Object identity}}\\[1pt]
\rotatebox{90}{\scriptsize\textbf{Wizard}}
& \includegraphics[width=\linewidth]{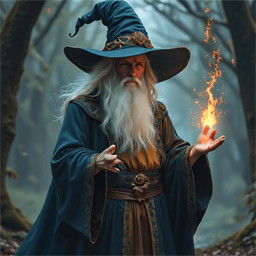}
& \includegraphics[width=\linewidth]{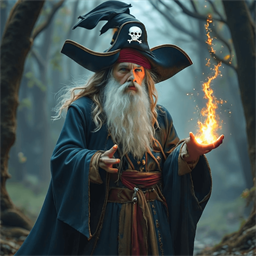}
& \includegraphics[width=\linewidth]{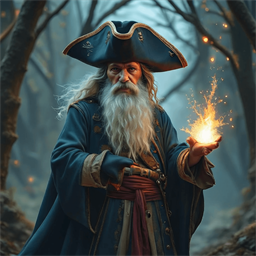}
& \includegraphics[width=\linewidth]{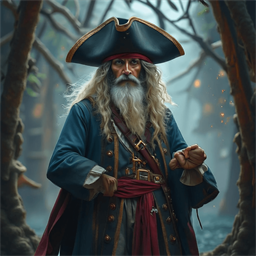}
& \includegraphics[width=\linewidth]{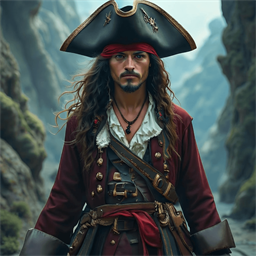}
& \includegraphics[width=\linewidth]{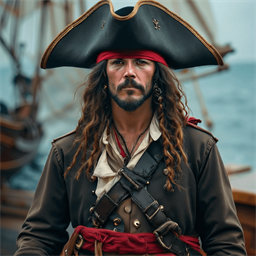}
& \includegraphics[width=\linewidth]{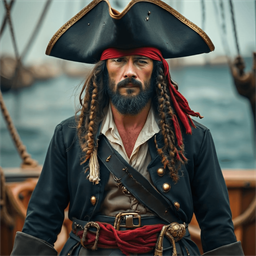}
\\[1pt]
\rotatebox{90}{\scriptsize\textbf{Monk}}
& \includegraphics[width=\linewidth]{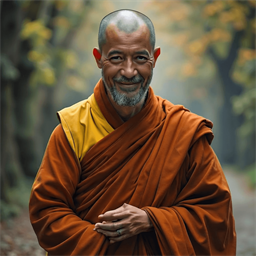}
& \includegraphics[width=\linewidth]{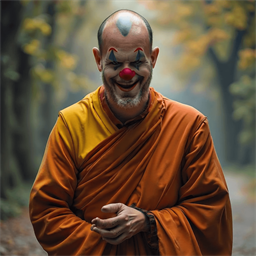}
& \includegraphics[width=\linewidth]{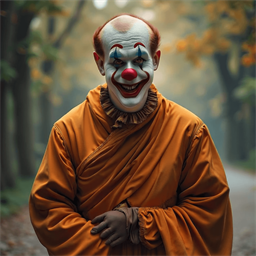}
& \includegraphics[width=\linewidth]{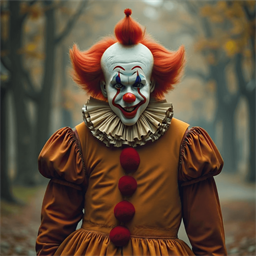}
& \includegraphics[width=\linewidth]{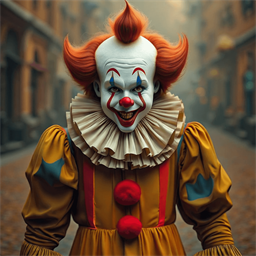}
& \includegraphics[width=\linewidth]{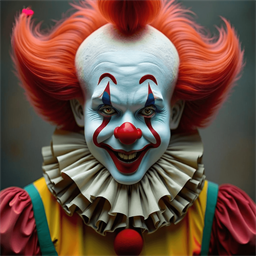}
& \includegraphics[width=\linewidth]{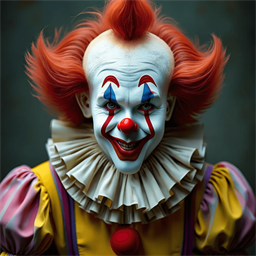}
\\[3pt]
\multicolumn{8}{@{}l}{\fontsize{7pt}{7pt}\selectfont\textbf{Texture / material}}\\[1pt]
\rotatebox{90}{\scriptsize\textbf{Bench}}
& \includegraphics[width=\linewidth]{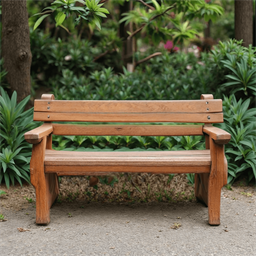}
& \includegraphics[width=\linewidth]{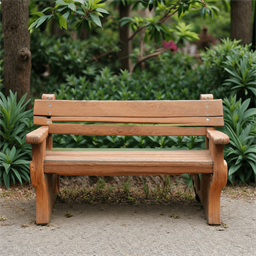}
& \includegraphics[width=\linewidth]{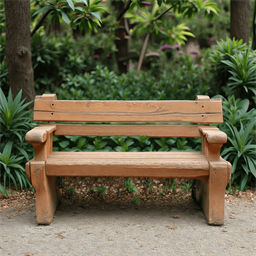}
& \includegraphics[width=\linewidth]{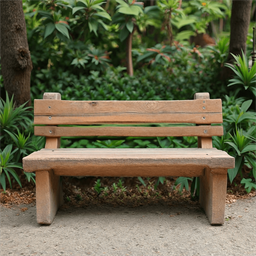}
& \includegraphics[width=\linewidth]{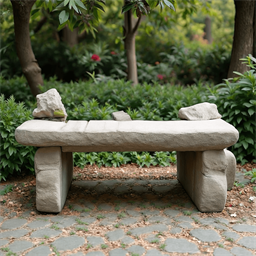}
& \includegraphics[width=\linewidth]{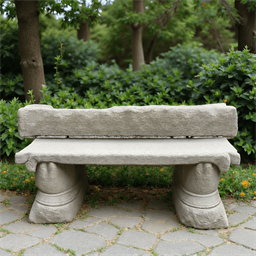}
& \includegraphics[width=\linewidth]{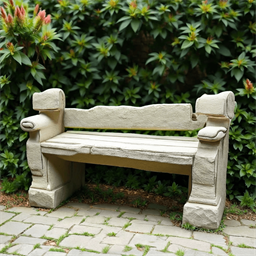}
\\[1pt]
\rotatebox{90}{\scriptsize\textbf{Spoon}}
& \includegraphics[width=\linewidth]{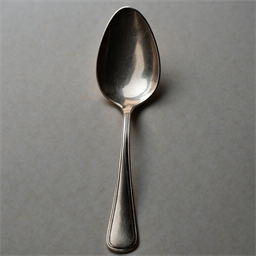}
& \includegraphics[width=\linewidth]{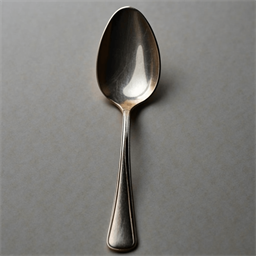}
& \includegraphics[width=\linewidth]{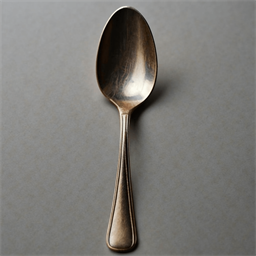}
& \includegraphics[width=\linewidth]{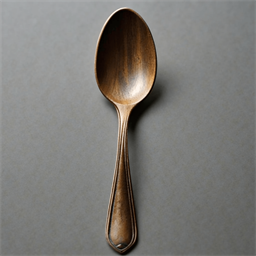}
& \includegraphics[width=\linewidth]{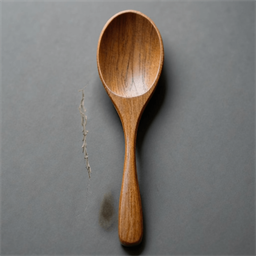}
& \includegraphics[width=\linewidth]{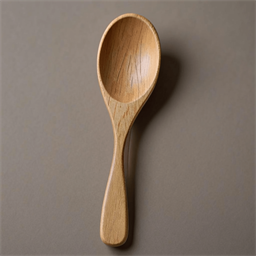}
& \includegraphics[width=\linewidth]{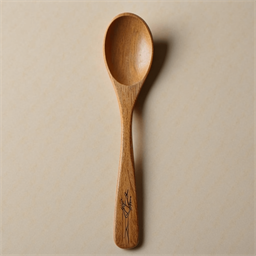}
\\
\end{tabular}
\end{minipage}
\caption{Two examples each of color, object-identity, and texture/material switches. Full A/B references bracket completed switched continuations. A column labeled s4, for example, uses A through scale 4 and B from scale 5 onward. Columns run from replacement after scale 4 to replacement after scale 0; earlier replacement increasingly reflects B.}
\label{fig:swap-qual}
\end{figure*}

We test prompt replacement after each completed scale from 0 to 5, with ten prompt pairs per property: color, object identity, style, and texture. Figure~\ref{fig:swap} compares the completed image with full-generation references for A and B; Figure~\ref{fig:swap-qual} shows six illustrative sequences.

\textbf{Replacing the prompt later leaves the final image closer to the original A reference.}
When we replace A immediately after scale 0 is generated, the completed image is more similar to B than to A for all four properties. When replacement occurs after scale 2, color and texture are already more similar to A. Object identity remains more similar to B when replacement occurs after scale 3, but is more similar to A when replacement is delayed until after scale 4. In the qualitative panel, full A/B generations bracket five completed switched continuations, ordered from late to early switches.

The reference-similarity ordering reverses between switches after scales 1 and 2 for color and texture, and after scales 3 and 4 for object identity; style is near balance after scale 1. These patterns suggest that several visual properties are already strongly constrained once scales 2--3 have been generated, while object identity remains responsive later. Early commitments persist through refinement, consistent with VISTA's emphasis on coarse-scale steering~\citep{shahabadi2026vista}. For search, this raises a practical opportunity: rank trajectories after these commitments begin to emerge, while substantial generation work remains.

\subsection{Selected attention losses anticipate final compositional alignment}

Early commitment makes selection valuable only if promising trajectories can be recognized. We therefore ask whether attention measured during generation predicts which candidate will better satisfy the same prompt at completion.

We compare three completed candidates per prompt using their earlier losses and final-image preferences. Per category, Infinity calibration uses 100 records and each backbone has 50 distinct held-out prompts, sharing text but not generations (Appendix~\ref{app:consensus}). Categories are color, shape, texture, 2D, and 3D; \emph{binding} groups the first three. Recorded scales are 2--8 (Appendix~\ref{app:protocol}).

A visual-language model judges entity existence and requested constraints in two reversed-order calls per pair. Answers are mapped back to candidate seeds. Agreement or one winner plus one tie yields a directional label; conflicting winners, two ties, and effectively tied losses are excluded. Appendix~\ref{app:protocol} specifies the judge, prompt, and tolerances.

Our primary metric asks how often retaining the lower-loss member of a pair would retain the preferred final image. For prompt $p$ and scale $s$, let $E_{p,s}$ contain evaluable pairs, $w_{pij}$ their observed winner, and $\widehat w_{pij,s}$ the loss-based prediction. Agreement over scale set $\mathcal S$ is
\begin{equation}
\operatorname{Acc}_{\mathrm{pair}}(\mathcal S)=
\frac{\sum_p\sum_{s\in\mathcal S}\sum_{(i,j)\in E_{p,s}}
\mathbf1[\widehat w_{pij,s}=w_{pij}]}
{\sum_p\sum_{s\in\mathcal S}|E_{p,s}|}.
\label{eq:pairwise-accuracy}
\end{equation}
Chance agreement is 50\%. The denominator counts pair--scale decisions: it does not average a candidate's losses across scales or simulate successive cuts.

Calibration examines 52 losses and 15 layer selections, then freezes a score and eight-layer block per category. Held-out testing uses the primary window 2--6. Appendix~\ref{app:protocol} gives the selection and aggregation rules.

\textbf{Frozen attention scores predict the preferred Infinity image most clearly for planar relations and color.}
Table~\ref{tab:atlas-heldout} reports 72.9\% agreement for planar relations and 62.7\% for color. Holm-adjusted permutation $p$-values are 0.0015 and 0.0308 across the original 15 locked endpoints. Texture reaches 60.3\% with adjusted $p=0.167$; shape and 3D provide less conclusive evidence. Thus predictive strength varies by constraint, even though all five point estimates exceed 50\%.

\textbf{Selected binding losses also transfer to Switti.}
Applying Infinity's frozen scores and layer windows to Switti yields 63.2\% agreement for color and 62.2\% for texture, supporting their predictive value on a second backbone (Table~\ref{tab:ranking-intervals}). These diagnostic comparisons reuse the selected windows. In the end-to-end experiments, Infinity uses those windows, whereas Switti uses all 30 layers. The latter setting tests ATLAS without imposing Infinity's restricted layer choices. Calibration estimates in Table~\ref{tab:atlas-heldout} remain subject to selection optimism.

\begin{table}[h]
  \caption{Pairwise agreement (\%) over indices 2--6. Both held-out sets use frozen Infinity choices. Chance is 50\%; indices need not denote matching resolutions. Intervals and prompt counts: Table~\ref{tab:ranking-intervals}.}
  \label{tab:atlas-heldout}
  \centering
  \small
  \setlength{\tabcolsep}{4pt}
  \begin{tabular}{lccc}
    \toprule
    Category & \Infinity{} calibration & \Infinity{} held-out & Switti held-out \\
    \midrule
    2D      & 68.6 & 72.9 & 56.7 \\
    Color   & 63.6 & 62.7 & 63.2 \\
    Shape   & 57.9 & 54.0 & 55.0 \\
    Texture & 65.6 & 60.3 & 62.2 \\
    3D      & 65.7 & 58.4 & 54.7 \\
    \bottomrule
  \end{tabular}
\end{table}

Pairwise agreement need not imply that the best of three candidates is selected. We therefore also test whether the unique lowest-loss candidate has the highest tie-aware preference score. Infinity reaches 61.7\% for 2D versus a 38.1\% random baseline, and 47.8\% for color versus 41.5\% (Table~\ref{tab:selection-audit}). This secondary test measures one decision among three completed paths, not survival through repeated cuts in a larger search.

\textbf{The attention signal strengthens after the earliest recorded scale.}
A score that becomes useful only at completion cannot save continuation work. Table~\ref{tab:scale-agreement} in Appendix~\ref{app:protocol} evaluates each recorded scale: Infinity's equal-category agreement rises from 55.7\% at scale 2 to 63.7\% at scale 3, stays near 62--64\% through scale 7, and reaches 57.9\% at scale 8. Useful ranking information is therefore available while finer-scale computation remains.

\subsection{Layer choice determines which attention an objective reads}

\begin{figure}[t]
  \centering
  \includegraphics[width=0.70\linewidth]{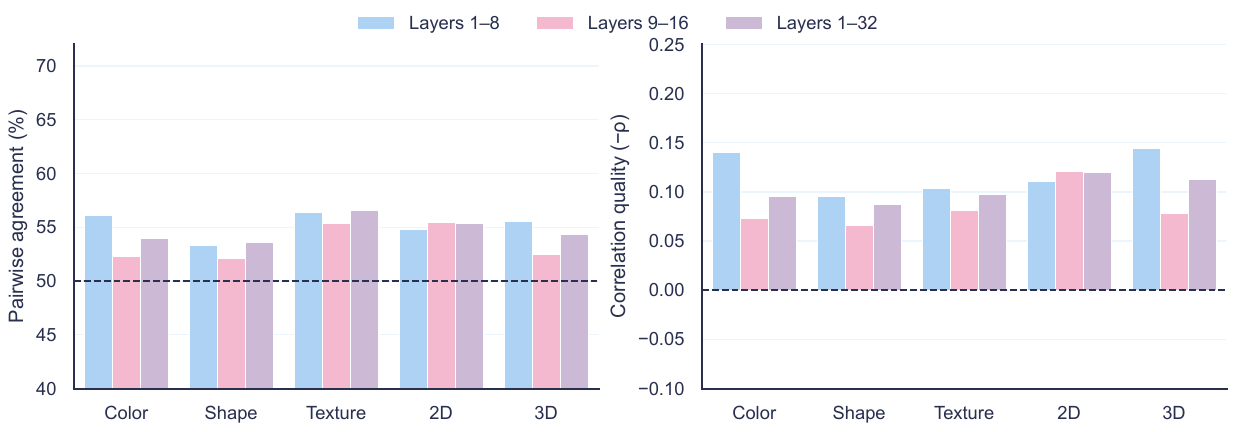}
  \caption{Comparing layers 1--8, 9--16, and all 32 on Infinity calibration prompts. Left: pairwise agreement, with the axis starting at 40\%; right: negative within-prompt Spearman correlation. Higher bars indicate better ranking. Dashed lines mark 50\% chance agreement and zero correlation. Appendix~\ref{app:protocol} defines aggregation across losses and scales.}
  \label{fig:atlas-dev-layers}
\end{figure}

Calibration compares 15 contiguous layer selections to identify which maps support ranking. Figure~\ref{fig:atlas-dev-layers} summarizes layers 1--8, 9--16, and all 32. The selected blocks are 1--8 for binding/depth and 9--16 for planar relations; depth broadens the best four-layer choice for simplicity (Appendix~\ref{app:protocol}).

\textbf{Fixed-loss comparisons retain much of the calibrated layer pattern.}
With each loss held fixed, the selected block has the highest held-out agreement in four Infinity categories; texture instead favors all layers (Table~\ref{tab:atlas-layer-transfer}). Switti point estimates likewise favor layers 1--8 for binding/depth and 9--16 for planar relations.

Agreement extends beyond the selected loss: Top-3 and Top-5 mean accuracies exceed 50\% by point estimate in every category (Table~\ref{tab:atlas-layer-transfer}). These averages of per-loss \emph{metrics} summarize several objectives; they are not candidate-score ensembles.

\begin{table*}[t]
  \centering
  \caption{Held-out agreement (\%) over 2--6. \emph{Top:} frozen losses, equally averaged over scales; blue marks selected blocks, All means 32/30 layers. Switti 3D coverage varies. \emph{Bottom:} Infinity pooled accuracy for Top-1 and per-loss metric averages for Top-3/Top-5. Aggregations differ; Top-5 is exploratory.}
  \label{tab:atlas-layer-transfer}
  \small
  \setlength{\tabcolsep}{4pt}
  \begin{tabular}{llcccccc}
    \toprule
    & & \multicolumn{3}{c}{\Infinity{}-2B} & \multicolumn{3}{c}{Switti} \\
    \cmidrule(lr){3-5}\cmidrule(lr){6-8}
    Category & Frozen block & 1--8 & 9--16 & All & 1--8 & 9--16 & All \\
    \midrule
    2D & 9--16
       & 64.1 & \textcolor{VISTABlue}{72.9} & 69.6
       & 47.9 & \textcolor{VISTABlue}{56.7} & 56.3 \\
    Color & 1--8
       & \textcolor{VISTABlue}{62.7} & 56.5 & 53.1
       & \textcolor{VISTABlue}{63.2} & 57.6 & 60.4 \\
    Shape & 1--8
       & \textcolor{VISTABlue}{54.0} & 45.1 & 49.1
       & \textcolor{VISTABlue}{55.0} & 43.1 & 44.7 \\
    Texture & 1--8
       & \textcolor{VISTABlue}{60.3} & 60.9 & 61.8
       & \textcolor{VISTABlue}{62.2} & 57.3 & 61.8 \\
    3D & 1--8
       & \textcolor{VISTABlue}{58.3} & 49.1 & 56.3
       & \textcolor{VISTABlue}{54.5} & 53.3 & 52.5 \\
    \bottomrule
  \end{tabular}
  \par\medskip
  \begin{tabular}{l>{\columncolor{SelectedColumnTint}}c
                    >{\columncolor{SelectedColumnTint}}c
                    >{\columncolor{SelectedColumnTint}}c}
    \toprule
    Category & Top-1 & Top-3 mean & Top-5 mean \\
    \midrule
    2D & 72.9 & 66.8 & 57.8 \\
    Color & 62.7 & 64.0 & 65.1 \\
    Shape & 54.0 & 55.6 & 53.7 \\
    Texture & 60.3 & 58.2 & 58.1 \\
    3D & 58.4 & 55.0 & 52.8 \\
    \bottomrule
  \end{tabular}
\end{table*}

\begin{figure}[t]
  \centering
  \includegraphics[width=0.70\linewidth]{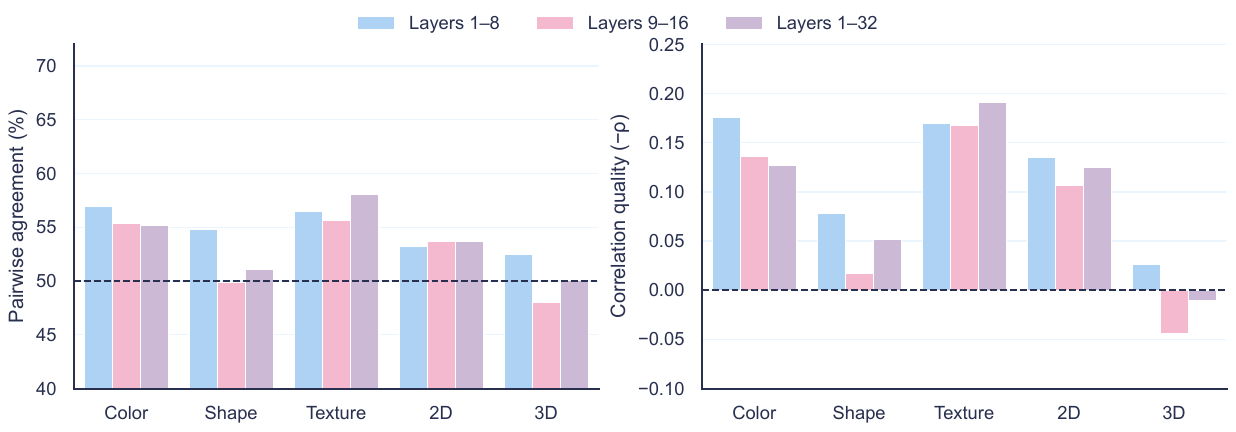}
  \caption{Exploratory Infinity held-out layer comparison using Figure~\ref{fig:atlas-dev-layers}'s aggregation. Agreement starts at 40\%; correlation is $-\rho$. Reference lines mark 50\% and zero. These post-lock aggregates do not redefine frozen choices.}
  \label{fig:atlas-heldout-layers}
\end{figure}

Layer preference also depends on the objective. Across the full loss bank, held-out averages favor early layers for color and all layers for texture; planar correlation favors early layers even though Separation ranks best with layers 9--16 (Figure~\ref{fig:atlas-heldout-layers}). The useful choice is therefore a loss--layer pairing, rather than a layer block alone.

Together, prompt switches reveal persistent early visual commitments, and within-prompt comparisons show that selected attention losses can rank trajectories before completion. The depth patterns are consistent with early attention organizing entity, attribute, and depth-related regions, while further processing benefits some planar scores. ATLAS turns this predictive information into a search policy, using attention to decide which paths receive the remaining computation.

\section{ATLAS: Attention-Guided Test-Time Latent Selection}
\label{sec:method}

\begin{figure}[h]
  \centering
  \includegraphics[width=0.66\linewidth]{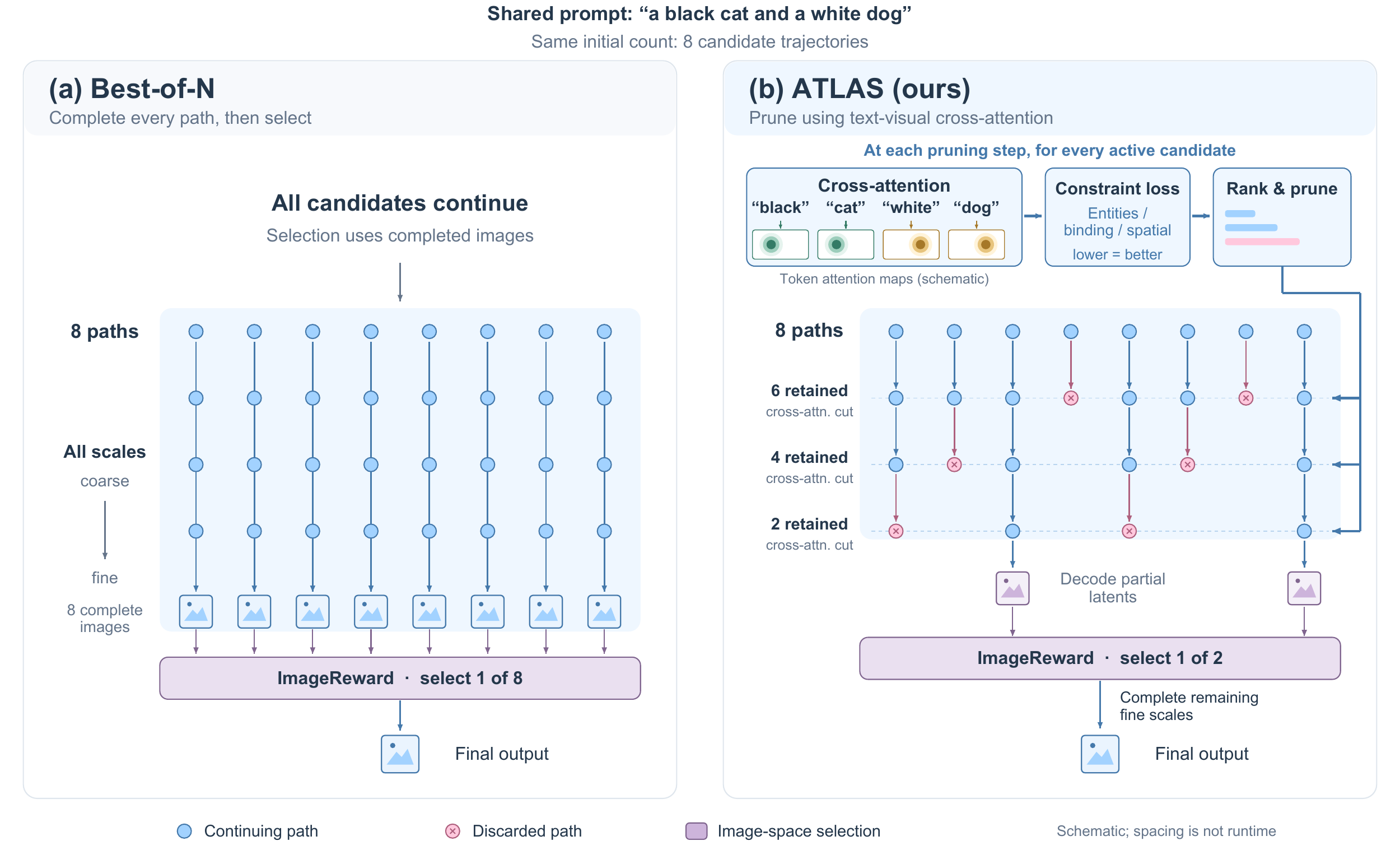}
  \caption{BoN and ATLAS at the same initial count. BoN completes all eight paths before ImageReward selection. ATLAS retains six, four, then two using losses computed from text--visual cross-attention; ImageReward selects one partial decode for completion. Crossed nodes stop. For larger ATLAS budgets, pre-reward counts scale by $m$. Icons and spacing are schematic, not image results or measured runtime.}
  \label{fig:atlas-method}
\end{figure}

Turning attention-based ranking into computational savings requires deciding which paths survive, how many remain, and when to commit to one. ATLAS uses category-specific scores for three cuts without decoding, a descending candidate budget, and one ImageReward decision to choose the path to complete (Figure~\ref{fig:atlas-method}).

\subsection{Frozen category scores rank active trajectories}

To identify promising trajectories, ATLAS applies attention losses matched to the prompt's binding and spatial constraints. Infinity uses the selected loss--layer pairs in Table~\ref{tab:atlas-frozen-scores}; Switti uses all 30 layers. Scores guide selection without changing parameters or surviving latents. SynGen and CONFORM follow prior objectives~\citep{rassin2023syngen,meral2024conform}; Appendix~\ref{app:loss-formulas} summarizes the objectives and key formulas.

A mixed-constraint prompt still requires one candidate ordering. Let $\mathcal D(p)$ index its available binding, planar, and depth terms. For candidate $k$ after scale $s$, ATLAS combines them as
\begin{equation}
  \ell_{k,s}(p)=\frac{1}{|\mathcal D(p)|}
    \sum_{d\in\mathcal D(p)}\loss_d(k,s;W_d),
  \qquad |\mathcal D(p)|>0.
  \label{eq:atlas-mixed-score}
\end{equation}
For prompts with multiple supported constraints, we average the corresponding losses and rank candidates by the resulting score, with lower values preferred. When no supported constraint is present, ATLAS remains inactive and generation is unchanged. Numeracy and non-spatial scores therefore retain the backbone values in our benchmark tables.

\begin{table}[h]
  \caption{Category-specific attention losses and selected transformer layers used by ATLAS on Infinity-2B. Layer indices are one-based.}
  \label{tab:atlas-frozen-scores}
  \centering
  \small
  \begin{tabular}{lll}
    \toprule
    Category & Attention loss & Infinity layers \\
    \midrule
    2D      & Separation (Appendix~\ref{app:loss-formulas}) & 9--16 \\
    Color   & SynGen~\citep{rassin2023syngen} & 1--8 \\
    Shape   & Peak (Appendix~\ref{app:loss-formulas}) & 1--8 \\
    Texture & CONFORM~\citep{meral2024conform} & 1--8 \\
    3D      & Footprint (Appendix~\ref{app:loss-formulas}) & 1--8 \\
    \bottomrule
  \end{tabular}
\end{table}

Ranking saves work by stopping lower-ranked paths before another scale. Let $\cand_s$ enter scale $s$, with $M_s=|\cand_s|$. At a scheduled attention cut after generating that scale, sort scores from lowest to highest, yielding identities $k_{(1)},\ldots,k_{(M_s)}$. If the next-scale budget is $K_{s+1}<M_s$, retain
\begin{equation}
  \cand_{s+1}
  =\{k_{(1)},\ldots,k_{(K_{s+1})}\}.
  \label{eq:atlas-retain}
\end{equation}
Retained paths continue from their existing latents; discarded paths stop. Scores determine which paths survive, while the schedule sets their number. All active paths continue between cuts. Appendix~\ref{app:controller-code} describes tie handling and skipped cuts.

\subsection{A descending budget concentrates computation on surviving paths}

Keeping many candidates supports exploration but increases continuation cost~\citep{chen2025ttsvar}. ATLAS preserves the initial pool through coarse scales, then reduces $8m\to6m\to4m\to2m\to1$, where $m$ is a positive integer budget multiplier. Attention controls the first three reductions; ImageReward makes the final choice.

Our benchmark label $N=m$ multiplies the base eight-path schedule: $8N$ initial paths narrow to $2N$ finalists, preserving the pruning proportions. Thus $N$ measures search-budget scaling, not candidate count. BoN reports completed candidates; cost profiles use actual initial count $n$ throughout.

Cut placement determines which later scales avoid processing discarded paths. With $K_s$ the target count \emph{entering} scale $s$, the schedules are
\begin{align}
\mathbf K^{\mathrm{Inf}}(m)
  &= (8m,8m,8m,8m,6m,4m,2m,2m,2m,1,1,1,1),
     \label{eq:atlas-infinity-budget}\\
\mathbf K^{\mathrm{Switti}}(m)
  &= (8m,8m,8m,8m,8m,8m,6m,4m,2m,2m,1,1,1,1).
     \label{eq:atlas-switti-budget}
\end{align}
Infinity cuts after completed scales $3,4,5$, and Switti after $5,6,7$. Their grids are $6^2,8^2,12^2$ and $7^2,9^2,12^2$, respectively, aligning cuts at nearby resolutions rather than identical indices. Pruning at these intermediate scales reduces the number of trajectories processed by the remaining finer scales. The remaining $2m$ paths continue to image-space selection.

\subsection{ImageReward selects one partial trajectory for completion}

The final choice evaluates rendered candidates beyond their token-map relationships. ATLAS decodes only the $2m$ finalists and scores prompt alignment with ImageReward~\citep{xu2023imagereward}. Let $D$ decode an accumulated prefix and $R_{\mathrm{IR}}$ score its prompt--image preference. After scale $s_R$, select the highest-reward candidate:
\begin{equation}
  k^\star\in\underset{k\in\cand_{s_R}}{\operatorname{argmax}}
  R_{\mathrm{IR}}\!\left(p,D(r_{k,\leq s_R})\right).
  \label{eq:atlas-imagereward}
\end{equation}
Here $s_R=8$ on Infinity and $s_R=9$ on Switti. Full-resolution decodes represent \emph{partial} latents, not completed generations; chunked decoding limits peak memory. Only the selected path completes the remaining scales. This combines attention-guided allocation with an image-space final choice, evaluated through output quality and inference cost rather than ranking agreement alone.

\section{Compositional Quality and Inference Cost}
\label{sec:evaluation}

We test whether ATLAS turns intermediate selection into better final composition and lower search cost. Benchmarks follow standard protocols with four seeds, except the single-run Infinity BoN baseline and random-pruning ablation. CompBench Avg.$_6$ covers binding, spatial relations, and complex composition; Overall is the mean across all eight categories, including numeracy and non-spatial relations. GenEval Avg. covers its three displayed categories.

\subsection{T2I-CompBench}

\paragraph{ATLAS improves targeted compositional scores as candidate budget grows.}
The backbone comparison tests whether more ATLAS computation improves composition (Table~\ref{tab:compbench-main}). Relative Avg.$_6$ gains are $4.3\%$, $7.3\%$, and $10.5\%$ on Infinity and $5.2\%$, $7.7\%$, and $10.6\%$ on Switti at budget multipliers $N=1,2,8$, using unrounded means. Complex composition improves at every budget; numeracy and non-spatial outputs remain unchanged. Both backbones therefore benefit from scaling this search policy. Appendix~\ref{app:four-seed-benchmarks} reports run-level variability.

\paragraph{ATLAS reaches competitive compositional quality across both backbones.}
TTS-VAR and BoN compare ATLAS with alternative uses of additional sampling. On Infinity, TTS-VAR has higher averages, while ATLAS leads on texture and complex composition at both shared budgets and on color at $N=2$. Switti ATLAS approaches BoN's averages at its largest setting, although ATLAS starts 64 paths and BoN completes eight. These are quality comparisons at the stated budgets; equal-count cost comparisons follow below. Infinity TTS-VAR $N=8$ exceeded A100 40 GB memory and is omitted. Published diffusion results provide context rather than matched controls~\citep{shahabadi2025infinitybeyond}.

\begin{table*}[h]
\centering
\caption{T2I-CompBench scores. Avg.$_6$ covers binding, spatial relations, and complex composition; Overall is the mean across all eight categories. Gray columns are untargeted and unchanged for ATLAS. ATLAS, TTS-VAR, and Switti BoN average four seeds; Infinity BoN uses one run. ATLAS and TTS-VAR start $8N$ paths; BoN completes $n=8$ candidates. Diffusion references:~\citet{shahabadi2025infinitybeyond}. Blue: ATLAS. Bold/underline: highest/second-highest per column across all models, including displayed ties.}
\label{tab:compbench-main}
\small
\setlength{\tabcolsep}{2.5pt}
\renewcommand{\arraystretch}{0.95}
\begin{tabular}{lcccccc>{\columncolor[gray]{0.92}}c>{\columncolor[gray]{0.92}}ccc}
\toprule
Model / setting & Color & Texture & Shape & 2D & 3D & Cmplx & Num. & Non-S. & Avg.$_6$ & Overall \\
\midrule
SDXL & 0.593 & 0.519 & 0.466 & 0.215 & 0.341 & 0.319 & \cellcolor[gray]{0.92}0.504 & \cellcolor[gray]{0.92}0.311 & 0.4088 & 0.4085 \\
PixArt-$\alpha$ & 0.407 & 0.444 & 0.367 & 0.202 & 0.350 & 0.324 & \cellcolor[gray]{0.92}0.506 & \cellcolor[gray]{0.92}0.308 & 0.3490 & 0.3635 \\
Flux-Dev & 0.746 & 0.644 & 0.482 & \underline{0.273} & 0.393 & 0.363 & \cellcolor[gray]{0.92}\textbf{0.613} & \cellcolor[gray]{0.92}0.309 & 0.4835 & 0.4779 \\
Flux-Schnell & 0.725 & 0.683 & 0.559 & 0.271 & 0.373 & 0.364 & \cellcolor[gray]{0.92}\underline{0.604} & \cellcolor[gray]{0.92}\underline{0.312} & 0.4958 & 0.4864 \\
\midrule
Switti & 0.732 & 0.620 & 0.500 & 0.208 & 0.378 & 0.366 & \cellcolor[gray]{0.92}0.555 & \cellcolor[gray]{0.92}\textbf{0.315} & 0.4673 & 0.4593 \\
\rowcolor{ATLASBlueTint}
\quad + \method{} ($N{=}1$, All) & 0.773 & 0.688 & 0.518 & 0.219 & 0.363 & 0.389 & \cellcolor[gray]{0.92}0.555 & \cellcolor[gray]{0.92}\textbf{0.315} & 0.4916 & 0.4774 \\
\rowcolor{ATLASBlueTint}
\quad + \method{} ($N{=}2$, All) & 0.799 & 0.695 & 0.524 & 0.227 & 0.387 & 0.390 & \cellcolor[gray]{0.92}0.555 & \cellcolor[gray]{0.92}\textbf{0.315} & 0.5035 & 0.4864 \\
\quad + BoN ($n{=}8$) & \underline{0.815} & \underline{0.708} & \textbf{0.568} & 0.226 & 0.400 & 0.399 & \cellcolor[gray]{0.92}0.555 & \cellcolor[gray]{0.92}\textbf{0.315} & \underline{0.5193} & 0.4983 \\
\rowcolor{ATLASBlueTint}
\quad + \method{} ($N{=}8$, All) & \textbf{0.817} & \underline{0.708} & 0.544 & 0.242 & 0.395 & 0.396 & \cellcolor[gray]{0.92}0.555 & \cellcolor[gray]{0.92}\textbf{0.315} & 0.5169 & 0.4964 \\
\midrule
\Infinity{}-2B & 0.749 & 0.632 & 0.475 & 0.234 & 0.405 & 0.384 & \cellcolor[gray]{0.92}0.572 & \cellcolor[gray]{0.92}0.310 & 0.4798 & 0.4701 \\
\quad + TTS-VAR ($N{=}1$) & 0.777 & 0.672 & 0.531 & 0.261 & 0.419 & 0.388 & \cellcolor[gray]{0.92}0.590 & \cellcolor[gray]{0.92}0.311 & 0.5080 & 0.4936 \\
\rowcolor{ATLASBlueTint}
\quad + \method{} ($N{=}1$) & 0.765 & 0.673 & 0.502 & 0.254 & 0.409 & 0.399 & \cellcolor[gray]{0.92}0.572 & \cellcolor[gray]{0.92}0.310 & 0.5004 & 0.4856 \\
\quad + TTS-VAR ($N{=}2$) & 0.783 & 0.682 & 0.549 & 0.264 & \textbf{0.425} & 0.393 & \cellcolor[gray]{0.92}\textbf{0.613} & \cellcolor[gray]{0.92}0.311 & 0.5159 & \underline{0.5024} \\
\rowcolor{ATLASBlueTint}
\quad + \method{} ($N{=}2$) & 0.785 & 0.695 & 0.523 & 0.262 & 0.421 & \underline{0.403} & \cellcolor[gray]{0.92}0.572 & \cellcolor[gray]{0.92}0.310 & 0.5147 & 0.4963 \\
\quad + BoN ($n{=}8$) & 0.739 & 0.614 & 0.456 & 0.233 & 0.398 & 0.383 & \cellcolor[gray]{0.92}0.565 & \cellcolor[gray]{0.92}0.309 & 0.4705 & 0.4621 \\
\rowcolor{ATLASBlueTint}
\quad + \method{} ($N{=}8$) & 0.801 & \textbf{0.716} & \underline{0.561} & \textbf{0.275} & \underline{0.424} & \textbf{0.405} & \cellcolor[gray]{0.92}0.572 & \cellcolor[gray]{0.92}0.310 & \textbf{0.5303} & \textbf{0.5080} \\
\bottomrule
\end{tabular}
\end{table*}

\paragraph{Attention-guided pruning improves composition, with a further gain from selected layers.}
Table~\ref{tab:compbench-layers} separates the scoring and layer choices while retaining ImageReward final selection. All-layer ATLAS improves Avg.$_6$ over random pruning by $2.45\%$ at $N=1$ and $4.27\%$ at $N=2$, using unrounded means. This comparison is descriptive: random pruning uses one seed, ATLAS four. Comparing the two attention settings on the same four seeds, selected layers add $0.5\%$ and $0.8\%$, despite mixed category changes. Switti's all-layer results show that restricted windows are not required for the framework.

\begin{table}[h]
\centering
\caption{Infinity pruning and layer ablations with ImageReward final selection. Random: one seed; attention: four-seed means. Avg.$_6$: displayed categories, excluding inactive Numeracy and Non-Spatial. Blue: selected-layer ATLAS. Bold/underline: highest/second-highest within each budget.}
\label{tab:compbench-layers}
\label{tab:compbench-random-pruning}
\scriptsize
\setlength{\tabcolsep}{3.5pt}
\begin{tabular}{llrrrrrrr}
\toprule
Pruning rule & Budget & Color & Texture & Shape & 2D & 3D & Cmplx & Avg.$_6$ \\
\midrule
Random & $N=1$ & 0.7510 & 0.6410 & 0.4820 & \underline{0.2450} & 0.4080 & 0.3890 & 0.4860 \\
Attention (All) & $N=1$ & \underline{0.7630} & \textbf{0.6748} & \underline{0.4985} & 0.2440 & \textbf{0.4107} & \underline{0.3965} & \underline{0.4979} \\
\rowcolor{ATLASBlueTint}
Attention (Selected) & $N=1$ & \textbf{0.7653} & \underline{0.6728} & \textbf{0.5023} & \textbf{0.2543} & \underline{0.4093} & \textbf{0.3988} & \textbf{0.5004} \\
\midrule
Random & $N=2$ & 0.7530 & 0.6430 & 0.4880 & 0.2510 & 0.4120 & 0.3920 & 0.4898 \\
Attention (All) & $N=2$ & \underline{0.7810} & \underline{0.6873} & \textbf{0.5235} & \underline{0.2578} & \underline{0.4140} & \underline{0.4010} & \underline{0.5108} \\
\rowcolor{ATLASBlueTint}
Attention (Selected) & $N=2$ & \textbf{0.7850} & \textbf{0.6945} & \underline{0.5233} & \textbf{0.2620} & \textbf{0.4208} & \textbf{0.4028} & \textbf{0.5147} \\
\bottomrule
\end{tabular}
\end{table}

\subsection{GenEval}

\paragraph{ATLAS also improves the targeted GenEval constraints on both backbones.}
GenEval tests whether the scaling trend extends to another benchmark (Table~\ref{tab:geneval}). The mean over color, color--attribute binding, and position rises at each budget on both backbones. Infinity's position scores increase throughout; Switti gains mainly in binding, with position improving at the largest budget.

\begin{table*}[h]
\centering
\caption{Targeted GenEval scores. Published diffusion and Infinity references:~\citet{shahabadi2025infinitybeyond}; ATLAS averages four seeds. Switti's parameter count:~\citet{voronov2025switti}. Blue: ATLAS. Bold/underline: highest/second-highest per score column across all models, including displayed ties.}
\label{tab:geneval}
\small
\setlength{\tabcolsep}{5pt}
\begin{tabular}{lccccc}
\toprule
\multirow{2}{*}{Method} & \multirow{2}{*}{\# Params} & \multicolumn{4}{c}{GenEval $\uparrow$} \\
\cmidrule(lr){3-6}
 & & Color & Color attr. & Position & Avg. \\
\midrule
\multicolumn{6}{l}{\emph{Diffusion models}} \\
SDXL & 3.5B & 0.862 & 0.210 & 0.105 & 0.392 \\
PixArt-$\alpha$ & 0.6B & 0.801 & 0.093 & 0.068 & 0.321 \\
Flux-Dev & 12B & 0.766 & 0.470 & 0.185 & 0.474 \\
Flux-Schnell & 12B & 0.785 & 0.505 & 0.263 & 0.518 \\
\midrule
\multicolumn{6}{l}{\emph{Autoregressive models}} \\
Switti & 2.5B & 0.880 & 0.328 & 0.145 & 0.451 \\
\rowcolor{ATLASBlueTint}
\quad + \method{} ($N{=}1$, All) & 2.5B & 0.891 & 0.415 & 0.145 & 0.484 \\
\rowcolor{ATLASBlueTint}
\quad + \method{} ($N{=}2$, All) & 2.5B & \underline{0.896} & 0.420 & 0.145 & 0.487 \\
\rowcolor{ATLASBlueTint}
\quad + \method{} ($N{=}8$, All) & 2.5B & \textbf{0.904} & 0.450 & 0.150 & 0.501 \\
\midrule
\Infinity{}-2B (published) & 2B & 0.830 & \underline{0.590} & 0.270 & 0.563 \\
\rowcolor{ATLASBlueTint}
\quad + \method{} ($N{=}1$) & 2B & 0.862 & \textbf{0.650} & 0.295 & 0.602 \\
\rowcolor{ATLASBlueTint}
\quad + \method{} ($N{=}2$) & 2B & 0.862 & \textbf{0.650} & \underline{0.310} & \underline{0.607} \\
\rowcolor{ATLASBlueTint}
\quad + \method{} ($N{=}8$) & 2B & 0.883 & \textbf{0.650} & \textbf{0.343} & \textbf{0.625} \\
\bottomrule
\end{tabular}
\end{table*}

\subsection{Inference Time and GPU Memory}

\paragraph{ATLAS reduces measured Infinity search cost at equal initial counts.}
To assess the computational benefit of early selection, Table~\ref{tab:atlas-efficiency} compares complete search procedures at equal initial counts on the same A100 40 GB, with 30 outputs per setting. At $n=8$ and $16$, Infinity ATLAS reduces generation time by $52.0\%$ and $70.8\%$ relative to BoN, and by $49.2\%$ and $60.9\%$ relative to TTS-VAR. Peak allocated memory is also lower than both baselines.

\paragraph{The time and memory savings extend to Switti.}
The Switti comparison tests whether savings extend beyond Infinity. Relative to BoN, ATLAS reduces time by $61.1\%$ and $73.1\%$ at $n=8$ and $16$, and peak allocated memory by $5.3\%$ and $13.1\%$. Both backbones thus show lower measured search cost at equal initial counts, rather than matched quality. Appendix~\ref{app:profiling-scope} details their scope.

\begin{table}[h]
\caption{A100 40 GB profiles at initial count $n$: 30 outputs per setting and mean s/output. Peak memory is rounded MiB; load allocation retains the source-reported MB label (Appendix~\ref{app:profiling-scope}). Blue: ATLAS. Bold/underline: lowest/second-lowest per metric within each backbone and initial count, excluding Base and including ties.}
\label{tab:atlas-efficiency}
\centering
\scriptsize
\setlength{\tabcolsep}{3.5pt}
\begin{tabular}{llrrrrr}
\toprule
Backbone & Method ($n$) & Outputs & Time (s) & \shortstack{Load alloc.\\(MB)} & \shortstack{Peak alloc.\\(MiB)} & \shortstack{Peak reserv.\\(MiB)} \\
\midrule
Switti & Base (1) & 30 & 1.34 & 6{,}720 & 8{,}732 & 15{,}620 \\
       & BoN (8) & 30 & \underline{5.24} & \textbf{8{,}474} & \underline{13{,}189} & \underline{20{,}472} \\
\rowcolor{ATLASBlueTint}
       & \method{} (8) & 30 & \textbf{2.04} & \textbf{8{,}474} & \textbf{12{,}485} & \textbf{18{,}200} \\
       & BoN (16) & 30 & \underline{9.54} & \textbf{8{,}474} & \underline{14{,}434} & \underline{22{,}158} \\
\rowcolor{ATLASBlueTint}
       & \method{} (16) & 30 & \textbf{2.57} & \textbf{8{,}474} & \textbf{12{,}537} & \textbf{18{,}200} \\
\midrule
\Infinity{}-2B & Base (1) & 30 & 2.63 & 7{,}342 & 11{,}520 & 22{,}178 \\
       & BoN (8) & 30 & 12.49 & \textbf{8{,}232} & 19{,}965 & 31{,}108 \\
       & TTS-VAR (8) & 30 & \underline{11.81} & \underline{8{,}819} & \underline{18{,}485} & \underline{24{,}466} \\
\rowcolor{ATLASBlueTint}
       & \method{} (8) & 30 & \textbf{6.00} & \textbf{8{,}232} & \textbf{16{,}414} & \textbf{21{,}914} \\
       & BoN (16) & 30 & 24.05 & \textbf{8{,}232} & 28{,}491 & 39{,}544 \\
       & TTS-VAR (16) & 30 & \underline{17.96} & \underline{8{,}819} & \underline{20{,}989} & \underline{27{,}726} \\
\rowcolor{ATLASBlueTint}
       & \method{} (16) & 30 & \textbf{7.02} & \textbf{8{,}232} & \textbf{17{,}416} & \textbf{24{,}774} \\
\bottomrule
\end{tabular}
\end{table}

\section{Limitations and Future Work}
\label{sec:limitations}

Broader prompts, human evaluation, and diversity audits would test how the measured gains extend beyond automated compositional scores. Adaptive selection rules could further balance exploration and continuation cost. Another direction is to combine ATLAS with VISTA's coarse-scale steering~\citep{shahabadi2026vista}: steering modifies developing trajectories, while ATLAS selects which receive further computation. Their complementary operations motivate testing a joint approach.

\section{Conclusion}

Internal attention can guide compositional search before final images are available. Within-prompt diagnostics identify predictive binding and spatial scores, with category-dependent transfer across backbones. ATLAS turns these on-the-fly signals into training-free selection, retaining promising trajectories while reducing the number that require further computation. Compositional benchmark means rise with search budget on both Infinity and Switti. Separate profiles show lower generation time and peak allocated memory than completing all candidates at equal initial counts. Together, these results connect internal compositional signals to scalable search: additional sampling expands exploration, while early selection limits continuation cost.

\label{sec:main-text-end}

\clearpage
\bibliography{main}
\bibliographystyle{tmlr}

\appendix

\section{Diagnostic Evaluation Protocol}
\label{app:protocol}

\paragraph{Records and settings.}
Infinity calibration has 100 records per category; each backbone has 50 held-out prompts per category, with seeds 1000--1002. Records join by prompt and seed. The 52 losses comprise nine planar, eleven per binding category, and ten depth definitions. Layers are one-based; recorded scales 2--8 are zero-based.

Reference settings use bfloat16 inference. Infinity-2B uses its associated visual tokenizer and Flan-T5-XL text encoder, guidance 3, temperature 0.5, top-$k$ 900, and top-$p$ 0.97. ScaleKV uses window/capacity/kernel $8/400/5$ with max pooling. Switti uses the 1024-resolution model, guidance 6, top-$k$ 400, top-$p$ 0.95, and disables guidance from index 10.

\paragraph{Preferences and exclusions.}
After deduplicating stored calls and mapping display positions to seeds, agreeing winners or a winner plus a tie define a directional pair. Conflicting winners and all-tie pairs are excluded. Missing matches and effective loss ties (relative tolerance $10^{-7}$, absolute tolerance $10^{-12}$) are also excluded. Signed 2D scores use absolute value; other frozen scores use identity.

Correlation and single-winner selection require a preference score for each candidate. Each candidate faces the other two candidates in both display orders, giving four judgments. We award one point for a win, half a point for a tie, and zero for a loss, then average these four values. Unlike pairwise agreement, this score retains the individual judgments even when reversing image order changes the winner.

At each prompt and scale, selection chooses the candidate with the lowest attention loss. We evaluate it when losses are available for all three candidates, the minimum loss is unique, and the VLM preference scores are not all equal. Selection is correct if the chosen candidate has the highest preference score, including a shared highest score. Its random baseline is the number of VLM-best candidates divided by three, averaged over eligible selections. In the tables, \emph{Top-1 loss} means the single objective chosen during calibration; \emph{selection accuracy} measures how often that objective chooses a VLM-best candidate.

\paragraph{Aggregation and calibration.}
Per-prompt Spearman and Kendall's $\tau_b$ require complete, nonconstant values. Prompt-centered Spearman subtracts each prompt's mean loss and preference before pooling candidates; figures display its negative. Layer figures average scale-wise metrics within windows 2--3, 2--4, 2--6, and 2--8, then equally across windows and losses. Early scales therefore recur. Primary agreement instead pools evaluable pair--scale decisions (Equation~\ref{eq:pairwise-accuracy}).

Calibration evaluates 15 layer selections and ranks losses by mean negative centered Spearman across those windows. The lock fixes loss order, an eight-layer block, and primary scales 2--6. Binding/depth use layers 1--8; planar relations use 9--16. Depth broadens the best four-layer block, 5--8, for simplicity. The original 15 held-out endpoints combine five categories with Top-1, mean Top-3 accuracy, and mean all-loss accuracy. The all-loss endpoint remains in multiplicity correction; Top-5 is later descriptive analysis.

\paragraph{Uncertainty and coverage.}
95\% intervals use the 2.5th/97.5th percentiles of 10,000 prompt-bootstrap replicates. Permutation tests reassign candidate identities within each prompt, sharing permutations across losses and scales. One-sided $p$-values use $(1+\#\{\mathrm{null}\geq\mathrm{observed}\})/10001$ from 10,000 replicates, followed by Holm correction across the original 15 endpoints. Marginal intervals and adjusted tests need not agree. Calibration intervals omit selection uncertainty; post-lock sweeps are descriptive. Switti's early-minus-middle contrasts remain uncertain: color $+5.6$ points, CI $[-4.6,16.0]$; texture $+4.9$, $[-3.1,13.0]$.

\begin{table}[h]
\centering
\small
\setlength{\tabcolsep}{4pt}
\caption{Infinity held-out coverage and secondary selection results for the frozen single loss over indices 2--6. $M$ counts evaluable pair--scale decisions; $U$ counts eligible prompt--scale selections. These repeated decisions are not independent samples. Selection and random baseline are percentages; selection intervals resample prompts. The adjusted $p$-value tests pairwise agreement within the original 15-endpoint family.}
\label{tab:selection-audit}
\begin{tabular}{lrrrlll}
\toprule
Category & $M$ & $U$ & Selection & 95\% CI & Random & $p_{\mathrm{Holm}}$ \\
\midrule
2D & 365 & 175 & 61.7 & [51.0, 72.1] & 38.1 & 0.0015 \\
Color & 405 & 205 & 47.8 & [38.2, 57.6] & 41.5 & 0.0308 \\
Shape & 350 & 190 & 38.4 & [27.8, 49.7] & 39.5 & 0.604 \\
Texture & 340 & 185 & 51.9 & [40.6, 62.7] & 39.6 & 0.167 \\
3D & 149 & 71 & 45.1 & [25.0, 65.8] & 37.6 & 0.604 \\
\bottomrule
\end{tabular}
\end{table}

\paragraph{Ranking intervals and scale coverage.}
\label{app:ranking-intervals}
Table~\ref{tab:ranking-intervals} gives prompt counts and intervals for Table~\ref{tab:atlas-heldout}. Table~\ref{tab:scale-agreement} averages category accuracies equally, bootstrapping prompts within categories. Final-image judgments recur across scales.

\begin{table}[h]
\centering
\small
\setlength{\tabcolsep}{4pt}
\caption{95\% prompt-bootstrap intervals for pairwise agreement (\%) and contributing prompt counts over indices 2--6. Held-out choices are frozen from Infinity calibration.}
\label{tab:ranking-intervals}
\begin{tabular}{lccc}
\toprule
Category & Infinity calibration & Infinity held-out & Switti held-out \\
\midrule
2D & \shortstack{[62.5, 74.7]\\$n=80$}
   & \shortstack{[64.5, 80.5]\\$n=35$}
   & \shortstack{[48.7, 64.5]\\$n=45$} \\
Color & \shortstack{[57.3, 69.4]\\$n=87$}
      & \shortstack{[54.4, 70.5]\\$n=42$}
      & \shortstack{[56.1, 70.0]\\$n=46$} \\
Shape & \shortstack{[51.7, 64.2]\\$n=75$}
      & \shortstack{[44.6, 62.8]\\$n=38$}
      & \shortstack{[44.7, 65.7]\\$n=39$} \\
Texture & \shortstack{[58.2, 72.8]\\$n=82$}
        & \shortstack{[51.2, 69.2]\\$n=37$}
        & \shortstack{[53.6, 70.2]\\$n=44$} \\
3D & \shortstack{[53.5, 76.8]\\$n=39$}
   & \shortstack{[43.9, 72.5]\\$n=23$}
   & \shortstack{[41.8, 66.9]\\$n=20$} \\
\bottomrule
\end{tabular}
\end{table}

\begin{table}[h]
\centering
\scriptsize
\setlength{\tabcolsep}{3pt}
\caption{Infinity held-out scale-wise pairwise agreement of frozen scores. Scale indices are zero-based; agreement and 95\% prompt-bootstrap intervals are percentages. Pair counts sum evaluable pairs across categories; multiple pairs can belong to one prompt. Chance agreement is 50\%.}
\label{tab:scale-agreement}
\begin{tabular}{lrrrrrrr}
\toprule
Scale & 2 & 3 & 4 & 5 & 6 & 7 & 8 \\
\midrule
Agreement & 55.7 & 63.7 & 62.0 & 63.7 & 63.1 & 62.3 & 57.9 \\
95\% CI & [49.0, 62.3] & [57.1, 70.2] & [56.1, 67.7] & [57.7, 69.5] & [57.0, 69.1] & [56.1, 68.3] & [51.1, 64.4] \\
Pairs & 318 & 324 & 323 & 322 & 322 & 320 & 319 \\
\bottomrule
\end{tabular}
\end{table}

\paragraph{Judge and comparison prompt.}
We use GPT-4o mini to compare completed images, presenting each pair twice in reversed order. Images are resized to at most 512 pixels. The judge returns a winner or a tie; an uninterpretable response is retried once and excluded if it remains unresolved. Candidate preference scores are calculated from these comparisons, not requested as separate scalar ratings.

The template below prioritizes entity existence, then annotated attributes or relations. For occlusion, existence requires recognizability rather than full visibility.
\begin{quote}\small
You are judging text-to-image results for compositional correctness.

Prompt: ``\texttt{\{prompt\}}''

Below are two images generated from this prompt, labelled Image 1 and Image 2.
Decide which one is BETTER using ONLY these criteria, in this order of importance:

1. Entity existence: \texttt{\{existence\}}

2. \texttt{\{category\}}. Specifically: \texttt{\{conditions\}}.

Ignore aesthetics, style, resolution and background details that the prompt does not mention.
If the two images are truly equal on both criteria, say so.

Reply with JSON only, no other text:
\begin{verbatim}
{"winner": 1, "reason": "one short sentence"}
\end{verbatim}
(use \texttt{"winner": 2} if Image 2 is better, or \texttt{"winner": 0} if they are equal)
\end{quote}

\section{Prompt-Switch Protocol and Scoring}
\label{app:switch-code}

Full A/B references share a seed. After scale $s$, the A prefix resumes under B with sampled maps, accumulated codes, next-scale input, self-attention cache, and sampler state preserved; only text conditioning and cross-attention tensors are rebuilt.

Scoring uses cosine similarities between normalized embeddings of completed switched and reference images. Means and marginal standard errors cover ten prompt pairs per property and scale. Crossover patterns are descriptive. The script supports CLIP, SigLIP, and DINOv2 with optional reference filtering; qualitative sequences are separate illustrations.

\section{Attention-Loss Definitions and Sources}
\label{app:loss-formulas}

This appendix provides context for the loss bank by summarizing objectives from prior work and the local variants evaluated in our study. SynGen motivates matching attributes to their nouns while separating unrelated pairs~\citep{rassin2023syngen}; CONFORM supplies a contrastive formulation~\citep{meral2024conform}; and Divide\&Bind supplies an attribute--noun binding objective~\citep{li2023divide}. SupCon draws on supervised contrastive learning~\citep{khosla2020supervised}. For spatial relations, PoS and PSE follow the probability-of-superiority approach of PSG~\citep{rezaei2025whysettle}, while the centroid implementation follows the spatial-violation construction of InfSplign~\citep{rastegar2025infsplign}. Separation is related to A-STAR's attention-segregation principle~\citep{agarwal2023astar}, and boundary containment follows VISTA's depth objective~\citep{shahabadi2026vista}. We summarize these objectives and their adaptations here, rather than introduce them as new contributions.

The bank contains eleven binding losses evaluated separately for color, shape, and texture, nine planar losses, and ten depth losses: 52 category-specific definitions in total. For layer-averaged attention $\bar A$ in Equation~\ref{eq:loss}, define phrase map $m_t(u)=|I_t|^{-1}\sum_{j\in I_t}\bar A(u,j)$ at visual position $u$, using valid token indices $I_t$. Objectives average over valid prompt constraints. Appendix~\ref{app:protocol} specifies the diagnostic sign convention.

\subsection{Selected objectives: Separation, Peak, and Footprint}

\paragraph{Separation.}
For valid noun pairs $\mathcal P$, cosine similarity penalizes shared attention support, not a specific spatial direction:
\begin{equation}
\ell_{\mathrm{sep}}=
\left|\frac{1}{|\mathcal P|}\sum_{(a,b)\in\mathcal P}
\operatorname{cos}(m_a,m_b)\right|.
\label{eq:app-separation}
\end{equation}
On nonnegative maps, the controller's negative cosine score gives the same ordering.

\paragraph{Peak.}
For the offline ranking study, we use integer grid coordinates. Let $c_t=\sum_u u\,\operatorname{softmax}(10m_t)_u$ and $D_{ij}=\|c_{\mathrm{adj},i}-c_{\mathrm{noun},j}\|_2^2$. For valid attribute--noun pairs $\mathcal B$ and other valid nouns $\mathcal J_i$,
\begin{equation}
\ell_{\mathrm{peak}}=
\frac{1}{|\mathcal B|}\sum_{i\in\mathcal B}
\left(D_{ii}-0.3\sum_{j\in\mathcal J_i}D_{ij}\right).
\label{eq:app-peak}
\end{equation}
Peak uses soft centroids. Competing nouns need not have valid attributes. In ATLAS, we use Peak for shape with coordinates normalized to $[0,1]$, sharpening factor 10, and competing-noun weight 0.3. Coordinate normalization preserves isolated rankings but changes mixed-loss scale.

\paragraph{Footprint.}
For valid depth relations $r=(a,b)\in\mathcal R$, set $p_t=\operatorname{softmax}(10m_t)$. In the offline ranking study, we use effective support $F_t=\exp[-\sum_u p_t(u)\log(p_t(u)+10^{-8})]$. Let $s_r=+1$ when $a$ is foreground relative to $b$, and $-1$ when background:
\begin{equation}
\ell_{\mathrm{foot}}=
\frac{1}{|\mathcal R|}\sum_{r=(a,b)\in\mathcal R}
\max\{0,s_r(F_b-F_a)\}.
\label{eq:app-footprint}
\end{equation}
Larger support is a depth proxy, also affected by object size and concentration. For ATLAS generation, we instead use coordinate spread:
$F_t=\sqrt{\operatorname{Var}_{p_t}(u_x)+10^{-6}}\sqrt{\operatorname{Var}_{p_t}(u_y)+10^{-6}}$ on $[0,1]^2$, with sharpening factor 10 and zero hinge margin. We retain this distinction between the diagnostic and generation objectives.

\subsection{Binding objectives}

Binding objectives measure whether an attribute and its noun attend to compatible regions. The local SynGen variant uses cosine similarity to favor matched pairs over unrelated nouns. KL and JSD compare their spatial attention distributions, while Divide\&Bind aligns those distributions after rescaling the attribute map. CONFORM treats matched attribute--noun embeddings as positives and other pairs as negatives; Exclusive and SupCon instead score the correct noun against competing nouns.

The remaining variants probe alternative matching statistics: Overlap measures Bhattacharyya overlap, Focal reweights cosine matching to emphasize poorly aligned pairs, and Peak compares soft centroids as defined above. The binding Composite equally averages the local SynGen and KL objectives. We evaluate these eleven losses separately on color, shape, and texture prompts. Our generation implementation uses SynGen with competing-noun weight 0.3 and CONFORM with temperature 0.07 and $3\times3$ Gaussian smoothing ($\sigma=0.5$). We include the objective implementations in the supplement.

\subsection{Planar objectives}

The nine planar objectives measure ordering, location, and separation. Centroid penalizes violations of the requested ordering between noun-map centers; Centroid-Prox adds similarity-based scoring for proximity relations. PoS and PSE use spatial distributions rather than centers to assess which noun lies on the requested side. Half-Field measures attention outside the requested image half, while Gaussian-Target compares it with a smooth target near the appropriate edge. Separation penalizes shared support. Our soft-box overlap objective, named IoU in the code, measures intersection area without dividing by union. Planar Composite combines separation and directional terms. Together, these variants test whether relative position or broader map geometry better predicts compositional preference.

\subsection{Depth objectives}

The ten depth objectives use attention geometry as a proxy for foreground--background ordering. Footprint and Entropy compare coordinate spread and effective support size, respectively. Depth-Peak favors a stronger foreground attention peak; Depth-Overlap favors foreground dominance where both maps have mass. Vertical PSE tests whether the foreground noun lies lower in the image, a ground-plane heuristic rather than a direct depth measurement.

Containment-Adjacency encourages the objects to be adjacent without merging their maps. Boundary Containment evaluates whether virtual points from the background boundary facing the foreground fall within foreground attention support. Depth Composite combines containment-adjacency, peak dominance, and overlap dominance. Separation and soft-box intersection complete the depth bank, providing relation-independent comparisons against these ordering-sensitive objectives.

\section{Controller Operation}
\label{app:controller-code}

\paragraph{Attention scoring.}
The controller averages text-conditioned attention across heads and selected layers before computing each loss. If the selected layers are unavailable, it uses all captured layers. Benchmark annotations identify the constraints; free-form parsing supports at most one relation. Each run uses one binding objective and equally averages it with applicable planar and depth losses, without rescaling their values. CONFORM requires at least two attribute--noun pairs; a configured alternative can handle fewer.

\paragraph{Pruning and final selection.}
At each scheduled pruning step, the controller retains the required number of lowest-loss candidates. Cutoff ties follow the sorting order, so the budget remains fixed; the offline diagnostic instead excludes tied comparisons. If attention maps or required loss terms are unavailable, the cut is skipped. The remaining finalists are decoded from their accumulated latents into full-resolution partial images, and ImageReward selects the trajectory to continue. Profiling decodes two finalists at a time to limit memory use.

\section{Data Integrity and Pair Consensus}
\label{app:consensus}

Each held-out category has 150 unordered pairs. Unanimous and winner-plus-tie pairs are directional; loss ties further reduce coverage.

\begin{table}[h]
\caption{Infinity held-out VLM consensus counts. Each row totals 150 unordered pairs. Subsequent effective loss ties further reduce the coverage of a given score.}
\label{tab:consensus}
\centering
\begin{tabular}{lrrrr}
\toprule
Category & Unanimous & Winner + tie & All tie & Conflict \\
\midrule
2D      & 47 & 26 & 45 & 32 \\
Color   & 44 & 37 & 51 & 18 \\
Shape   & 41 & 29 & 54 & 26 \\
Texture & 32 & 36 & 49 & 33 \\
3D      & 52 & 25 & 28 & 45 \\
\bottomrule
\end{tabular}
\end{table}

Audits verify complete calls, exact order reversal, consistent duplicate records, and answer-to-seed mapping. Reversal balances positions, not all judge biases.

Trimmed exact texts show no calibration/held-out overlap; both held-out backbones share 50 distinct prompts per category. Calibration has 100 distinct texts except shape/depth, with 99 each. Their duplicate texts remain separate record-ID bootstrap clusters, so calibration intervals omit this dependence as well as selection uncertainty.

\section{Reproducibility and Data Provenance}

We release our analysis and generation code, frozen configurations, diagnostic records, and benchmark, prompt-switch, and profiling summaries. We document model identifiers, weight-hash prefixes or snapshot revisions, environments, and launch scripts. For CompBench, we use the category-specific losses in Table~\ref{tab:atlas-frozen-scores}; for GenEval, we use SynGen for binding and Separation for position. Layer selections and budgets follow Section~\ref{sec:method}; Appendices~\ref{app:four-seed-benchmarks} and~\ref{app:profiling-scope} specify benchmark seeds and profiling settings.

Our extraction script restores diagnostic JSON files with checksum verification, and the README lists reproduction commands. We provide manuscript sources and qualitative figures separately. Generation requires the upstream models, weights, and evaluation dependencies. We anonymize release copies without changing measurements. We calculate CompBench means and sample standard deviations from valid category values, excluding missing or erroneous cells.

\section{Multi-Seed Benchmark Reporting}
\label{app:four-seed-benchmarks}

We use seeds 1001--1004 for standard CompBench and GenEval runs, except the single-run Infinity BoN baseline and random-pruning ablation. Table~\ref{tab:compbench-seeds} reports per-run means and sample standard deviations; Run 1--4 index the four repetitions. We calculate Avg.$_6$ and Overall within each run before averaging and report category means for GenEval. The random-pruning ablation covers six targeted categories at $N=1,2$. We include benchmark summaries and ablation configurations in the supplement.

\begin{table}[h]
\caption{T2I-CompBench Avg.$_6$ for four repetitions, with means and sample standard deviations for Avg.$_6$ and Overall (the eight-category mean). ATLAS $N=8$ begins with 64 trajectories, while BoN $n=8$ completes eight.}
\label{tab:compbench-seeds}
\centering
\small
\setlength{\tabcolsep}{4pt}
\begin{tabular}{lcccccc}
\toprule
Backbone / setting & Run 1 & Run 2 & Run 3 & Run 4 & Avg.$_6\pm$ SD & Overall $\pm$ SD \\
\midrule
\Infinity{}-2B / TTS-VAR $N{=}1$ & 0.5150 & 0.5075 & 0.5035 & 0.5058 & $0.5080\pm0.0050$ & $0.4936\pm0.0042$ \\
\Infinity{}-2B / TTS-VAR $N{=}2$ & 0.5130 & 0.5177 & 0.5170 & 0.5160 & $0.5159\pm0.0021$ & $0.5024\pm0.0014$ \\
\Infinity{}-2B / \method{} $N{=}1$ & 0.5010 & 0.4963 & 0.5037 & 0.5007 & $0.5004\pm0.0030$ & $0.4856\pm0.0023$ \\
\Infinity{}-2B / \method{} $N{=}2$ & 0.5177 & 0.5138 & 0.5127 & 0.5147 & $0.5147\pm0.0021$ & $0.4963\pm0.0016$ \\
\Infinity{}-2B / \method{} $N{=}8$ & 0.5292 & 0.5273 & 0.5340 & 0.5307 & $0.5303\pm0.0028$ & $0.5080\pm0.0021$ \\
Switti / \method{} $N{=}1$ & 0.4912 & 0.4927 & 0.4915 & 0.4910 & $0.4916\pm0.0008$ & $0.4774\pm0.0006$ \\
Switti / \method{} $N{=}2$ & 0.5087 & 0.4983 & 0.5012 & 0.5058 & $0.5035\pm0.0046$ & $0.4864\pm0.0035$ \\
Switti / \method{} $N{=}8$ & 0.5232 & 0.5145 & 0.5123 & 0.5177 & $0.5169\pm0.0047$ & $0.4964\pm0.0035$ \\
Switti / BoN $n{=}8$ & 0.5242 & 0.5160 & 0.5163 & 0.5208 & $0.5193\pm0.0039$ & $0.4983\pm0.0029$ \\
\bottomrule
\end{tabular}
\end{table}

\section{Inference Profiling Protocol}
\label{app:profiling-scope}

We profile equal initial candidate counts on an A100 40 GB, with 30 outputs per setting. We use color prompts, seed 1000, one discarded warm-up, two-candidate decoding batches, and expandable memory segments for every method. We keep Infinity's text encoder on the GPU. Our timing code synchronizes CUDA before and after each generation call and resets peak-memory counters before timing. Initial model loading, attention-hook setup, and image-file writing lie outside that interval. We report mean generation time and maximum allocated and reserved memory across outputs, dividing peak bytes by $2^{20}$ (MiB). We also report model-load allocation, labeled MB in our profiling records.

\section{Broader Impact}

More efficient compositional generation can support visual content creation, but can also facilitate misleading imagery. Selection may amplify biases in the generator or reward model. Deployment should retain appropriate content safeguards and evaluate demographic bias and output diversity; compositional benchmark scores alone do not assess these risks.

\end{document}